# Identifying Mentions of Pain in Mental Health Records Text: A Natural Language Processing Approach

Jaya Chaturvedi (*Institute of Psychiatry, Psychology and Neurosciences, King's College London*),
Sumithra Velupillai (*Institute of Psychiatry, Psychology and Neurosciences, King's College London*),
Robert Stewart (*Institute of Psychiatry, Psychology and Neurosciences, King's College London, Health Data Research UK, South London and Maudsley Biomedical Research Centre, London, United Kingdom*)
Angus Roberts (*Institute of Psychiatry, Psychology and Neurosciences, King's College London, Health Data Research UK*)

Pain is a common reason for accessing healthcare resources and is a growing area of research, especially in its overlap with mental health. Mental health electronic health records are a good data source to study this overlap. However, much information on pain is held in the free text of these records, where mentions of pain present a unique natural language processing problem due to its ambiguous nature. This project uses data from an anonymised mental health electronic health records database. The data are used to train a machine learning based classification algorithm to classify sentences as discussing patient pain or not. This will facilitate the extraction of relevant pain information from large databases, and the use of such outputs for further studies on pain and mental health. 1,985 documents were manually triple-annotated for creation of gold standard training data, which was used to train three commonly used classification algorithms. The best performing model achieved an F1-score of 0.98 (95% CI 0.98-0.99).

## 1. Introduction

Pain is defined as an unpleasant sensory and emotional experience, and is influenced by a variety of biological, psychological, and social factors [1]. Pain is a common reason for people to access healthcare facilities, thereby making electronic health records (EHR) a potential source for information on pain [2].

EHRs are longitudinal compilations of electronic data pertaining to a person's medical history or healthcare [3]. They have been increasingly used in research as they provide the opportunity to explore patient symptoms and findings from structured and unstructured fields. Since pain is not well recorded in these structured fields, it may help to supplement this information with data from unstructured clinical text [4].

A commonly used machine learning based NLP approach is text classification, in which labels are assigned to units of text (sentences/paragraphs/documents) [5]. Commonly used classification algorithms include Support Vector Machines [6–8] and K-Nearest Neighbours [9–11]. Recent state of the art approaches use embedding models and transformer-based neural network architectures [12], such as the bi-directional encoder representations of BERT [13]. Many healthcare domain related models have emerged, such as PubMedBERT [14], BioBERT [15], ClinicalBERT [16], UmlsBERT [17] and SAPBERT [18] which were developed after recognition of the need for specialized models due to linguistic differences between general and biomedical text [19].

This paper describes the methods undertaken to develop an NLP application for a sentence-level classification of mentions of physical pain within clinical text. Two BERT models were trained - bert_base and SAPBERT - and compared to two conventional models - support vector machines (SVM) and K-Nearest Neighbours (KNN). To the best of our knowledge, such extraction of information about pain from mental health clinical text using NLP has not been done.

## 2. Methods

### 2.1. Data Source

An anonymised version of EHR data from The South London and Maudsley NHS Foundation Trust (SLaM), one of the largest mental healthcare organizations in Europe, is stored in the Clinical Record Interactive Search

(CRIS) database [20]. The infrastructure of CRIS has been described in detail with an overview of the cohort profile [21]. CRIS contains over 30 million documents, averaging 90 documents per patient [22]. There are 23 different text sources (such as attachments, event notes, nurse assessment letters, etc.). Most of the text is contained within attachments and event notes, and so these were used as the data sources in this project.

*2.2. Ethics and Data Access*

Ethics approval for CRIS has been granted by (Oxford C Research Ethics Committee, reference 18/SC/0372). Research projects that use the CRIS database are reviewed and approved by a patient-led oversight committee (described in [23]). An opt-out model is in place for service users and is advertised in all publicity material and initiatives. Data are owned by a third party, Maudsley Biomedical Research Centre (BRC), who run the CRIS tool, providing access to anonymised data. These data can only be accessed by permitted individuals from within a secure firewall.

*2.3. Data Extraction*

Pain can be described in numerous ways, using a variety of terms. To help identify which documents in CRIS might be discussing pain, a lexicon of such pain terms was developed from a combination of pain-related terms extracted from the literature and biomedical ontologies, supplemented with additional similar terms from word embedding models. This lexicon and its development has been described in more detail in [24]. Terms from this pain lexicon were used to identify documents within CRIS that might be discussing pain. Documents containing pain terms were extracted for further processing using SQL. No time or diagnosis filter was applied to the extraction.

*2.4. Annotation Task*

Extracted documents were used to create a corpus of text discussing patient pain by labelling, i.e., annotating spans of text as being about pain or not. Each span consisted of 200 characters before and after a pain-related term. First, a set of annotation guidelines were developed to provide rules defining when a sentence should be considered as discussing pain. Next, terms from the pain lexicon were highlighted in the extracted documents. Three medical student annotators read through the extracted documents considering these spans of text containing the previously highlighted pain terms. Annotators labelled each span with one of three labels: *relevant* i.e., referring to physical pain experienced by the patient; *not relevant* i.e., mentions not related to pain, not related to the patient or hypothetical and metaphorical mentions; and *negated* i.e., absence of pain. Inter-annotator agreements were calculated after multiple rounds where 200 documents were annotated by all three annotators in each round. The annotation tool used for this was MedCAT [25]. Any disagreements were discussed, and the annotation guidelines updated. This iterative process was carried out until an inter-annotator agreement of over 0.80 was achieved, after which each annotator was then given a separate set of documents. The documents that were annotated by all three annotators during the iteration process were adjudicated following a set of adjudication guidelines in line with the most recent version of annotation guidelines. The annotation and adjudication guidelines can be accessed online[1].

*2.5. NLP application*

The annotations were split into train/test/validation sets at a proportion of 80/10/10 respectively. Four different models were trained, as detailed in Table 1. The parameters were chosen based on the recommendations made in (21) and models were checked for overfitting.

Table 1. Model specifications

| Model | Tokenizer | Pre-processing | Other Parameters |
|---|---|---|---|
| 1. Support Vector Machine | NLTK | Lowercase, stopword, white space and punctuation removal, lemmatize and tokenize | Tf-Idf vectorizer Default parameters from sklearn |
| 2. K-Nearest Neighbour | | | |
| 3. BERT | bert_base_uncased | Tokenize Prepend sentence with special token [CLS] and append with special token [SEP] | Epochs: 3 Batch size: 16 Optimizer: AdamW, learning rate 3e-5 |
| 4. SAPBERT | cambridgeltl/SapBE | | Epochs: 4 |

---

[1] https://github.com/jayachaturvedi/pain_in_mental_health/blob/main/Annotation%20Guidelines%20-%20Pain%20-%20for%20github.pdf

|  |  |  |
|---|---|---|
| RT-from-PubMedBERT-fulltext | Pad and truncate sentence to max length 105 (default is 511) | Batch size: 16<br>Optimizer: AdamW, learning rate 2e-5 |

## 3. Results

### 3.1. Data Extraction

A total of 1,985 randomly selected documents from 723 patients were extracted that contained pain related keywords from the lexicon. The most common diagnosis codes for these extracted patients were Mood disorders (ICD10 chapters F30-39) (33% of patients). There was an average of 8 annotations per patient.

### 3.2. Annotations

An inter-annotator agreement of 90% (Cohen's kappa 0.88) was achieved after four rounds (each round containing 200 documents) of triple annotations. A total of 5,644 annotations were obtained. 72% of these were marked as relevant, 15% as not-relevant, and 13% as negated. The relevant annotations were labelled as 1. The not-relevant and negated annotations were combined and labelled as 0. 71% annotations were labelled as class 1 (relevant) for both training and testing data.

### 3.3. Evaluation of NLP application

A single GPU (Tesla T4) was used for the training models. K-fold validation was carried out for evaluation of the models, and 95% confidence intervals were calculated. The result for each algorithm is outlined in Table 2. The BERT models performed better than Support Vector machine and K-Nearest Neighbour models.

Table 2. Evaluation Metrics, including 95% confidence intervals

| Model | Precision | Recall | F1-score (average from 10-fold cross validation) |
|---|---|---|---|
| Support Vector Machine | 0.86 (0.83-0.88) | 0.98 (0.97-0.99) | 0.91 (0.90-0.93) |
| K-Nearest Neighbour | 0.84 (0.81-0.87) | 0.91 (0.89-0.93) | 0.87 (0.85-0.89) |
| BERT | 0.96 (0.94-0.97) | 0.98 (0.97-0.99) | 0.97 (0.96-0.98) |
| SAPBERT | 0.98 (0.97-0.99) | 0.99 (0.98-0.99) | 0.98 (0.98-0.99) |

### 3.4. Error Analysis

During the annotation process, common disagreements included when an instance could be interpreted as physical or metaphorical, such as "...causing him pain", and hypothetical mentions such as "...she feared the pain" and "?migraine".

After training the models, some false positives spotted during error analysis on the test data. For the BERT_base model, there were instances such as "...wishing to project his pain on others", "father's hip pain". Some false negatives such as "denying symptoms other than stomach ache", "...if pain increases" were also noted.

The SAPBERT model showed false negatives when there were undecipherable symbols incorporated in the text, which might have occurred during the anonymisation process of the text, as well as misspellings or conjoined words such as "dabdominal pain" and "achespainodd sensations". False positives were instances such as "risk of potential pressure sores".

## 4. Discussion

The ambiguous nature of pain was highlighted during this project, especially during the annotation process where it took multiple rounds for three clinically trained annotators to agree on the meanings and interpretations of the pain mentions. Bearing this in mind, it is understandable that the classification models struggled with hypothetical and metaphorical instances. This highlights the importance of context and the necessity for the NLP models to incorporate and consider context during the classification task. This is a strength of transformer-based models such as BERT, which could be why they performed better than SVM/KNN.

Amongst the two BERT models that were trained, SAPBERT, which was pre-trained using a biomedical ontology, UMLS, performed slightly better than bert_base. There were differences in how each of the BERT models used in this project tokenised words, where SAPBERT was able to tokenise clinical concepts more accurately. This improvement in tokenisation might have impacted and improved the overall performance of the model.

## 5.  Conclusions

The objective of this project was to develop a machine learning based NLP application that can classify mentions of pain within clinical text as relevant or not. BERT models, which use a transformer-based machine learning technique and contextual embeddings, outperformed the other algorithms. This is a novel approach towards extracting information about pain from mental health records, leveraging the unstructured clinical notes to identify patients with relevant mentions of pain, and such cohorts of patients can then further be used in epidemiological and other pain related research with more confidence in the actual occurrence of pain when mentioned in the text.

**Acknowledgements**

This work uses data provided by patients and collected by the NHS as part of their care and support. The authors are also grateful to Dr Aurelie Mascio for providing access to some of her Python scripts.